\documentclass[lettersize,journal]{IEEEtran}
\usepackage[utf8]{inputenc}
\usepackage[T1]{fontenc}
\usepackage{amsmath,amsfonts}
\usepackage{algorithmic}
\usepackage{algorithm}
\usepackage{array}
\usepackage[caption=false,font=normalsize,labelfont=sf,textfont=sf]{subfig}
\usepackage{textcomp}
\usepackage{stfloats}
\usepackage{url}
\usepackage{verbatim}
\usepackage{graphicx}
\usepackage{cite}
\usepackage{booktabs}
\usepackage{siunitx}
\usepackage{multirow}
\usepackage{colortbl}
\usepackage{enumitem}
\usepackage[skip=0.333\baselineskip]{caption} 
\usepackage[table]{xcolor} 
\usepackage{hyperref}
\begin{document}

\title{RadioFlow: Efficient Radio Map Construction Framework with Flow Matching}

\author{Jia~Haozhe\IEEEauthorrefmark{1},
        Wenshuo~Chen\IEEEauthorrefmark{1},
        Xiucheng~Wang,
        Nan~Cheng,
        Hongbo~Zhang,
        Kuimou~Yu,
        Songning~Lai,
        Nanjian~Jia,
        Bowen~Tian,
        Hongru~Xiao,
        and~Yutao~Yue\IEEEauthorrefmark{2}%
\thanks{\IEEEauthorrefmark{1} Equal contribution.}%
\thanks{\IEEEauthorrefmark{2} Corresponding author: Yutao Yue (e-mail: yutaoyue@hkust-gz.edu.cn).}%
\thanks{Jia Haozhe, Wenshuo Chen, Kuimou Yu, Songning Lai, Bowen Tian, and Yutao Yue are with Hong Kong University of Science and Technology (Guangzhou), Guangzhou 511453, China.}%
\thanks{Xiucheng Wang and Nan Cheng are with the State Key Laboratory of ISN and School of Telecommunications Engineering, Xidian University, Xi'an 710071, China.}%
\thanks{Nanjian Jia is with Peking University, Beijing 100871, China.}%
\thanks{Hongbo Zhang is with Beijing Technology and Business University, Beijing 100048, China.}%
\thanks{Hongru Xiao is with the College of Civil Engineering, Tongji University, Shanghai 200092, China.}%
\thanks{Institute of Deep Perception Technology, JITRI, Wuxi 214000, China.}%
} 

\markboth{RadioFlow: Efficient Radio Map Construction Framework with Flow Matching}%
{Haozhe et al.: RadioFlow: Efficient Radio Map Construction Framework with Flow Matching}

\maketitle

\begin{abstract}
Accurate and real-time radio map (RM) generation is crucial for next-generation wireless systems, yet diffusion-based approaches often suffer from large model sizes, slow iterative denoising, and high inference latency, which hinder practical deployment. To overcome these limitations, we propose \textbf{RadioFlow}, a novel flow-matching-based generative framework that achieves high-fidelity RM generation through single-step efficient sampling. Unlike conventional diffusion models, RadioFlow learns continuous transport trajectories between noise and data, enabling both training and inference to be significantly accelerated while preserving reconstruction accuracy. Comprehensive experiments demonstrate that RadioFlow achieves state-of-the-art performance with \textbf{up to 8$\times$ fewer parameters} and \textbf{over 4$\times$ faster inference} compared to the leading diffusion-based baseline (RadioDiff). This advancement provides a promising pathway toward scalable, energy-efficient, and real-time electromagnetic digital twins for future 6G networks. We release the code at \href{https://github.com/Hxxxz0/RadioFlow}{GitHub}.
\end{abstract}


\section{Introduction}


The rapid evolution of wireless communication systems toward 5G Advanced and 6G is driving unprecedented demands for accurate environment awareness to support intelligent network planning, dynamic resource management, and adaptive service provisioning. Future networks must operate under stringent performance requirements, including ultra high data rates, ultra low latency, massive connectivity, and high reliability across diverse and dynamic environments such as dense urban areas, complex indoor layouts, and air space ground integrated networks. Achieving these goals requires a comprehensive understanding of the spatial distribution of electromagnetic (EM) propagation characteristics, including signal strength, interference, and pathloss, to enable proactive optimization and robust decision-making. To meet this need, radio maps, also known as radio environment maps (REM),\cite{9931518, pesko2014radio}have emerged as a critical technology, providing fine grained, location specific EM information over large geographical areas. These maps serve as foundational enablers for a wide range of next generation applications, from intelligent beamforming and user localization to electromagnetic digital twins and environment aware AI driven networking, making their accurate and efficient construction a central challenge in modern wireless communications.

Despite recent progress, the construction of high-fidelity radio maps remains a fundamentally challenging task, particularly in scenarios demanding real time adaptability and large scale deployment. Traditional measurement-based approaches though empirically reliable are resource intensive, time consuming, and inherently unscalable, making them unsuitable for rapidly changing environments or large geographical areas. Alternatively, ray tracing simulations \cite{7152831,9811058} provide fine-grained physical modeling but incur prohibitive computational overhead and lack the responsiveness required for dynamic wireless systems \cite{9811058}. To address these limitations, machine learning based methods have been proposed, yet they present critical tradeoffs between accuracy, efficiency, and generalizability. For instance, convolutional architectures such as RadioUNet \cite{9354041,10693745,9747240,10014763} achieve fast inference but often produce over-smoothed radio maps with limited spatial detail, diminishing their utility in applications that require fine-grained propagation structure. In contrast, diffusion based generative models like RadioDiff \cite{10764739} achieve state of the art fidelity by capturing complex signal variations, yet are severely constrained by iterative denoising procedures, resulting in long inference times that are incompatible with real time 6G use cases. Further, advanced generative methods such as Transformer based RadioNet \cite{9753644,10627516,7869126,10946503,10082228} and GAN based RME-GAN \cite{9662616,10130091} suffer from training instability, poor generalization to unseen environments, and significant simulation to reality gaps, especially when trained on limited or synthetic data. These challenges collectively underscore the urgent need for a new class of generative models that can offer both high-fidelity and fast inference, while maintaining robustness across diverse and dynamic wireless scenarios.

To address these critical limitations, we propose a novel generative modeling framework called RadioFlow, leveraging Flow Matching (FM) based on Continuous Normalizing Flows (CNFs). Flow Matching \cite{lipman2023flowmatchinggenerativemodeling} operates without dependence on simulations, directly regressing conditional vector fields to map noise to the target radio map distribution. This innovative approach enables one-step, high-quality map generation, overcoming the slow iterative denoising of diffusion models. A key advantage of FM is its flexibility in accommodating diverse probability paths, such as those inspired by Optimal Transport (OT), which enhances robustness in modeling complex environments. Our proposed method, RadioFlow, harnesses this by utilizing interpolation trajectories conditioned explicitly on environmental scenes (e.g., urban layouts). We then leverage an Ordinary Differential Equation (ODE) formulation to deterministically transport probability mass along these paths. This ODE-based approach not only enables stable, single-step sampling but also preserves probability mass, a principle analogous to energy conservation in electromagnetic propagation, thus ensuring high-quality generation. Importantly, we implement RadioFlow with a simplified yet highly effective UNet architecture, demonstrating that state-of-the-art performance can be achieved without the overhead of intricate designs. The main contributions of this paper are summarized as follows.
\begin{enumerate}
    \item We propose RadioFlow, the first radio map generation framework based on Flow Matching with CNFs. Unlike traditional diffusion-based models, RadioFlow enables single-step high-fidelity generation by directly learning conditional vector fields, eliminating the need for iterative denoising.
    \item We design a scene-aware conditioning mechanism that guides the generative process using environmental context (e.g., urban layouts), allowing the model to adaptively capture location-dependent propagation characteristics and improve spatial accuracy.
   \item We develop a lightweight UNet-based architecture enhanced with conditional embeddings, achieving state-of-the-art accuracy. The Lite variant reduces parameter count by about 8.3× and inference latency by 9.5× compared to diffusion-based baselines such as RadioDiff, while the Full variant achieves 4.6× faster inference with higher reconstruction fidelity.

\end{enumerate}

\begin{figure*}[t]
    \centering
    \includegraphics[width=\textwidth, keepaspectratio]{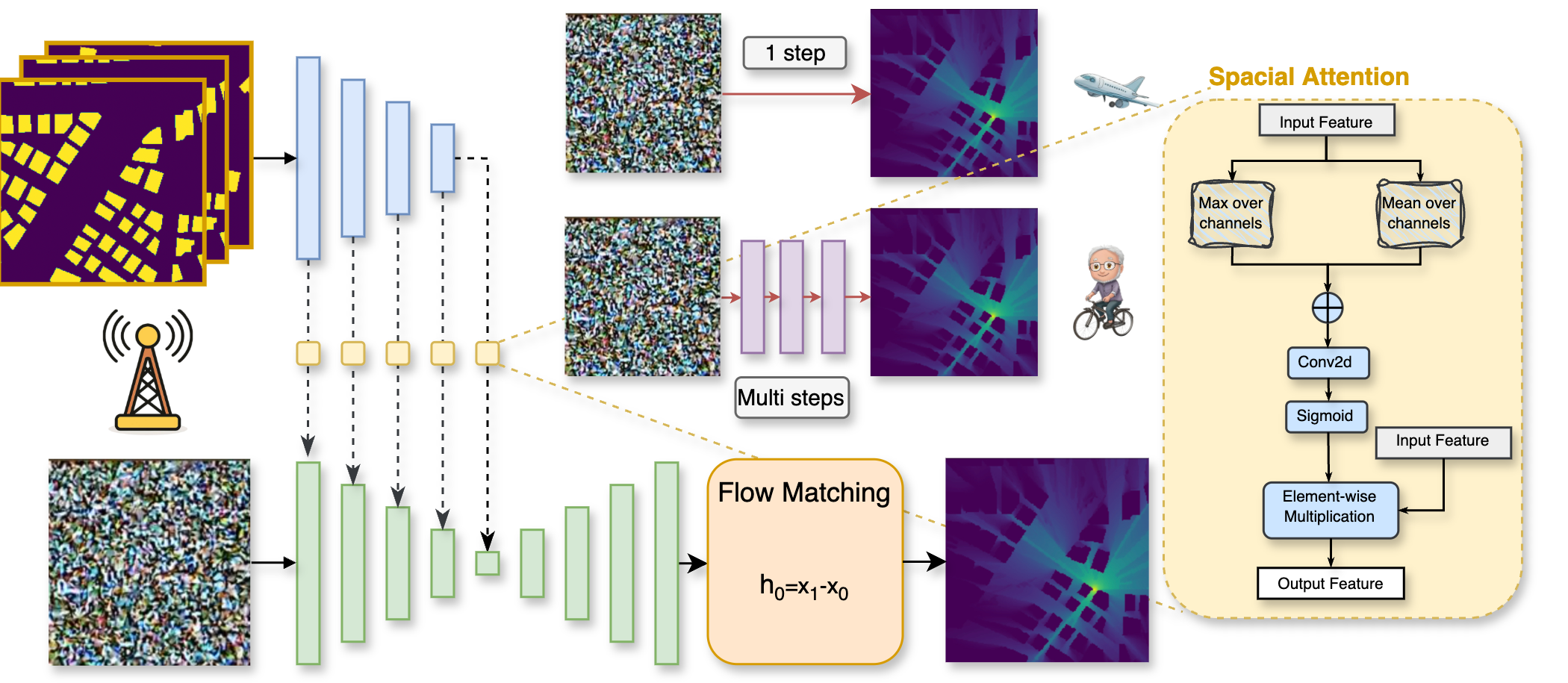}
    \caption{
    Overview of the proposed RadioFlow architecture. The model leverages Conditional Flow Matching (CFM) to learn deterministic vector fields that map noise to the target radio map in a single step. A UNet-based encoder-decoder processes the input, and a spatial attention module refines the output by emphasizing important propagation regions.
    }
    \label{fig:model_architecture}
\end{figure*}


\section{Related Work}

Traditional model-free approaches for radio map estimation have gained popularity by leveraging deep learning techniques to predict signal strength directly from data, bypassing the need for explicit physical models. One prominent method is RadioUNet \cite{9354041,10693745,9747240,10014763}, which uses a UNet-based convolutional neural network (CNN) to estimate pathloss based on urban geometry and transmitter locations. This model achieves a high accuracy with a root mean square error (RMSE) of approximately 1 dB, while also offering rapid computation times between $10^{-3}$ to $10^{-2}$ seconds. RadioUNet has shown significant improvements over traditional ray-tracing methods, offering a more scalable and efficient approach. However, it still faces limitations in capturing detailed local variations and complex propagation environments. 

Similarly, RME-GAN \cite{9662616,10130091} employs a conditional generative adversarial network (cGAN) with a two-phase learning framework to reconstruct detailed radio maps from sparse measurements. RME-GAN effectively captures global propagation patterns as well as local environmental effects, making it well-suited for scenarios with limited measurement data. Despite its strengths, RME-GAN struggles to handle dynamic or highly variable environments and faces challenges in maintaining training stability in the presence of complex environmental factors.

In recent years, diffusion-based models have emerged as a powerful alternative for generating high-fidelity radio maps, particularly for complex and dynamic environments. RadioDiff \cite{10764739} introduces a denoising diffusion probabilistic model that eliminates the need for explicit sampling. By integrating an attention-enhanced U-Net with adaptive fast Fourier transform (FFT) modules, RadioDiff excels at capturing high-frequency details, outperforming previous methods in terms of metrics like normalized mean square error (NMSE) and structural similarity index (SSIM) on the RadioMapSeer dataset. It often surpasses RadioUNet by $10$–$15\%$ in dynamic environments, offering superior flexibility and adaptability. However, the computational demands of iterative denoising processes inherent to diffusion models make them unsuitable for real-time applications, especially in the context of rapidly changing environments such as those found in 5G and 6G systems.

To address the high computational cost and slow inference speeds of diffusion models \cite{10419041,10081412}, the Radio Map Diffusion Model (RMDM) \cite{jia2025rmdmradiomapdiffusion} combines physics-informed neural networks (PINNs) with a dual U-Net diffusion framework. RMDM incorporates physical constraints such as the Helmholtz equation, ensuring the generated radio maps adhere to known physical laws. This hybrid approach yields impressive results in terms of NMSE (0.0031 in static scenarios) and shows superior generalization when dealing with sparse data. However, despite its improvements in computational efficiency, RMDM still faces challenges in maintaining real-time performance, particularly in dynamic and large-scale environments. 

In summary, while existing methods such as RadioUNet, RME-GAN, and diffusion-based models like RadioDiff and RMDM have made significant strides in radio map generation, they still face critical limitations, particularly when it comes to real-time applications. These include issues with computational overhead, training instability, poor generalization to unseen environments, and reliance on simulated datasets. The need for efficient, high-quality radio map generation remains a critical challenge for next-generation wireless communication systems, including 5G-Advanced and 6G networks.




\section{Method}

\subsection{Preliminaries}

Flow Matching (FM) methods \cite{lipman2023flowmatchinggenerativemodeling} constitute a class of simulation-free generative modeling frameworks that directly parameterize and learn continuous normalizing flows through deterministic vector fields. Unlike traditional diffusion models, which iteratively denoise data samples via stochastic Markov chains, Flow Matching establishes deterministic conditional vector fields that facilitate efficient and exact transport between predefined probability distributions.

Formally, let $q(\mathbf{x}_1)$ represent the data distribution defined on $\mathbb{R}^d$, and let $p_0(\mathbf{x}_0) = \mathcal{N}(\mathbf{0}, \mathbf{I})$ denote a standard Gaussian prior distribution. FM constructs conditional probability paths $p_t(\mathbf{x}|\mathbf{x}_0, \mathbf{x}_1)$ that smoothly interpolate between initial Gaussian states $\mathbf{x}_0 \sim p_0(\mathbf{x}_0)$ and target data points $\mathbf{x}_1 \sim q(\mathbf{x}_1)$. In practice, these intermediate probability paths at time $t \in [0,1]$ can be explicitly defined by the interpolation:
\begin{equation}
    \mathbf{x}_t = (1 - t)\mathbf{x}_0 + t \mathbf{x}_1 + \sigma_t \boldsymbol{\epsilon}, \quad \boldsymbol{\epsilon} \sim \mathcal{N}(\mathbf{0}, \mathbf{I})
\end{equation}
where $\sigma_t \geq 0$ controls the variance of intermediate distributions, and often $\sigma_t$ can be chosen to vanish or follow a deterministic schedule. In a typical scenario, choosing $\sigma_t = 0$ yields deterministic straight-line paths, simplifying analysis and optimization.

Flow Matching leverages conditional vector fields, which directly encode the transport direction and magnitude between paired states $(\mathbf{x}_0, \mathbf{x}_1)$:
\begin{equation}
    \mathbf{u}_t(\mathbf{x}|\mathbf{x}_0, \mathbf{x}_1) = \mathbf{x}_1 - \mathbf{x}_0
\end{equation}
These conditional fields enable closed-form evaluation of gradients, explicitly providing the conditional score function:
\begin{equation}
    \nabla \log p_t(\mathbf{x}|\mathbf{x}_0, \mathbf{x}_1) = -\frac{\mathbf{x} - (1 - t)\mathbf{x}_0 - t \mathbf{x}_1}{\sigma_t^2}
\end{equation}
when $\sigma_t > 0$. In the deterministic limit ($\sigma_t \rightarrow 0$), this expression simplifies, reinforcing the deterministic nature of transport.

The global vector field utilized by the continuous normalizing flow is then obtained by marginalizing over data and initial Gaussian distributions:
\begin{equation}
    \mathbf{u}_t(\mathbf{x}) = \mathbb{E}_{q(\mathbf{x}_1), p_0(\mathbf{x}_0)} \left[ \frac{p_t(\mathbf{x}|\mathbf{x}_0, \mathbf{x}_1)}{p_t(\mathbf{x})} \mathbf{u}_t(\mathbf{x}|\mathbf{x}_0, \mathbf{x}_1) \right]
\end{equation}
where $p_t(\mathbf{x}) = \int q(\mathbf{x}_1)p_0(\mathbf{x}_0)p_t(\mathbf{x}|\mathbf{x}_0, \mathbf{x}_1) d\mathbf{x}_0 d\mathbf{x}_1$ is the marginal probability at time $t$. To avoid explicit computation of these marginals, FM methods minimize the following conditional regression objective:
\begin{equation}
    \mathcal{L}_{\text{FM}}(\theta) = \mathbb{E}_{t, q(\mathbf{x}_1), p_0(\mathbf{x}_0), p_t(\mathbf{x}|\mathbf{x}_0,\mathbf{x}_1)} \left[ \| \mathbf{v}_{\theta}(t,\mathbf{x}) - \mathbf{u}_t(\mathbf{x}|\mathbf{x}_0, \mathbf{x}_1) \|^2 \right]
\end{equation}
Here, $\mathbf{v}_{\theta}(t,\mathbf{x})$ represents the neural network parameterization approximating the conditional vector field.

Flow Matching offers significant theoretical and practical advantages over diffusion-based methods. Notably, the deterministic nature of the conditional transport paths minimizes the Wasserstein-2 distance between distributions, guaranteeing geometric optimality. Consequently, FM does not require iterative denoising steps. Instead, sampling is performed by numerically solving a single ordinary differential equation (ODE):
\begin{equation}
    d\mathbf{x} = \mathbf{v}_\theta(t,\mathbf{x}) dt,
\end{equation}
often achievable with as few as one Euler discretization step, drastically reducing computational complexity while maintaining high-quality sample generation. This reduction from hundreds of steps to a single-step generation represents a critical efficiency advancement over diffusion models. Furthermore, the deterministic coupling ensures exact adherence to the continuity equation without stochastic approximation errors, promoting stable and accelerated training.

In summary, Flow Matching's combination of explicit conditional paths, deterministic vector fields, and optimal transport properties offers a robust, computationally efficient, and theoretically elegant approach to generative modeling, significantly improving upon traditional diffusion methodologies.
\subsection{ODE vs SDE}
\paragraph{(1) Continuity Equation and Energy Conservation.}
The ODE-based formulation in Flow Matching naturally satisfies the \textit{continuity equation} that governs the temporal evolution of the probability density:
\begin{equation}
\frac{\partial p_t(\mathbf{x})}{\partial t} + \nabla_\mathbf{x} \cdot \big(\mathbf{v}_\theta(\mathbf{x}, t)\, p_t(\mathbf{x})\big) = 0,
\label{eq:continuity}
\end{equation}
where $p_t(\mathbf{x})$ denotes the probability density of the generated signal at time $t$, and $\mathbf{v}_\theta(\mathbf{x}, t)$ represents the learned deterministic velocity field.  
Equation~\eqref{eq:continuity} enforces \textbf{probability mass conservation}, i.e.,
the total probability $\int p_t(\mathbf{x})\,d\mathbf{x}$ remains invariant over time. 
In other words, the generative process transports the existing probability mass smoothly across space without creation or annihilation.

In the context of radio map generation, Eq.~\eqref{eq:continuity} is mathematically analogous to the \textbf{Poynting theorem} in electromagnetism, which expresses energy conservation of the electromagnetic field:
\begin{equation}
\frac{\partial u(\mathbf{x},t)}{\partial t} + \nabla_\mathbf{x} \cdot \mathbf{S}(\mathbf{x},t) = 0,
\label{eq:poynting}
\end{equation}
where $u(\mathbf{x},t)$ is the electromagnetic energy density and $\mathbf{S}(\mathbf{x},t) = \mathbf{E} \times \mathbf{H}$ denotes the Poynting vector (energy flux).  
Both Eq.~\eqref{eq:continuity} and Eq.~\eqref{eq:poynting} share the same divergence-conservation structure, ensuring that the total ``mass'' (probability or energy) is conserved and only redistributed spatially. Compared with the deterministic ODE formulation, 
SDE-based diffusion models introduce stochastic perturbations that inherently violate the conservation property, leading to non-physical fluctuations in the generated radio signal distributions. Specifically, the diffusion process can be described by a stochastic differential equation (SDE):
\begin{equation}
d \mathbf{x}_t = \mathbf{f}_\theta(\mathbf{x}_t, t)\, dt + g(t)\, d\mathbf{W}_t,
\label{eq:sde}
\end{equation}
where $\mathbf{f}_\theta(\mathbf{x}_t, t)$ denotes the drift term, $g(t)$ is the diffusion coefficient,
and $d\mathbf{W}_t$ represents the standard Wiener process (Gaussian white noise).  
The corresponding evolution of the probability density $p_t(\mathbf{x})$ follows the \textit{Fokker--Planck equation (FPE)}:
\begin{equation}
\frac{\partial p_t(\mathbf{x})}{\partial t}
= - \nabla_\mathbf{x} \cdot \big( \mathbf{f}_\theta(\mathbf{x}, t) p_t(\mathbf{x}) \big)
+ \frac{1}{2} \nabla_\mathbf{x}^2 \big( g^2(t) p_t(\mathbf{x}) \big).
\label{eq:fpe}
\end{equation}
The first term represents deterministic advection,
while the second diffusion term
$\frac{1}{2}\nabla_\mathbf{x}^2(g^2(t)p_t(\mathbf{x}))$
models the stochastic spreading of the density induced by noise.

Therefore, the ODE formulation inherently preserves \textbf{signal power continuity and spatial smoothness}, aligning with the deterministic nature of electromagnetic propagation.  
In contrast, SDE-based diffusion models introduce stochastic noise terms that violate such conservation and often lead to non-physical fluctuations in the generated radio fields.

\paragraph{(2) Energy-Minimal Transport under Optimal Transport Theory.}
From the perspective of optimal transport, Flow Matching minimizes the \textit{Wasserstein-2 distance} between the source and target distributions by optimizing the expected kinetic energy of the learned flow field:
\begin{equation}
\min_{\mathbf{v}_\theta} \int_0^1 \mathbb{E}_{\mathbf{x}_t} \big[\|\mathbf{v}_\theta(\mathbf{x}_t, t)\|^2 \big] \, dt.
\label{eq:ot}
\end{equation}
This objective corresponds to finding the \textbf{energy-minimal transport path} that continuously moves samples from the noise distribution to the target radio map distribution.  
Physically, this principle mirrors the \textbf{least energy dissipation law} governing electromagnetic field propagation, where energy travels along the path minimizing total action or transmission loss.

Consequently, the ODE formulation enforces a deterministic, energy-efficient mapping consistent with real-world radio physics---preserving both energy conservation and minimal-dissipation flow while maintaining global spatial coherence.  
This property is particularly desirable for modeling radio maps, where the underlying field variations are governed by deterministic electromagnetic interactions rather than stochastic diffusion.

\subsection{Model Architecture}

The RadioFlow framework proposes a novel generative architecture that synthesizes high-fidelity radio maps through streamlined flow-based transformations. As illustrated in Figure~1, the architecture operates through three coordinated stages: input conditioning, flow-based generation, and contextual refinement.\\
\textbf{Input Conditioning.} The model accepts two inputs: (1) an environmental or contextual map representing the desired signal distribution, and (2) a Gaussian noise image serving as the generation seed; the target radio map is used only during training as ground-truth supervision. These inputs are connected through learned interpolation trajectories that establish a continuous transformation pathway. The trajectories are governed by conditional probability paths $p_t(\mathbf{x} \mid \mathbf{x}_0, \mathbf{x}_1)$, where parameter $t \in [0,1]$ controls the transition from noise ($t=0$) to target ($t=1$).\\
\textbf{Flow-Based Generation.} At the architecture's core lies a flow-matching module that enables single-step synthesis through deterministic vector field learning. Unlike iterative denoising approaches, this module directly learns a conditional vector field $\mathbf{u}_t(\mathbf{x} \mid \mathbf{x}_0, \mathbf{x}_1, \mathbf{c})$ that maps the noise distribution to the target radio map distribution, where the conditioning is explicitly provided through environmental context $\mathbf{c}$. The transformation is achieved by solving an ordinary differential equation parameterized by a neural network, significantly accelerating generation speed while maintaining spatial coherence.\\
\textbf{Contextual Refinement.} A spatial attention module \cite{woo2018cbamconvolutionalblockattention} enhances the initial radio map estimate by incorporating environmental context. This module processes auxiliary features (e.g., terrain maps, building layouts) through multi-scale feature extraction and channel-wise aggregation. Spatial attention weights are then computed to emphasize propagation-critical regions, followed by feature recalibration to produce the final enhanced radio map. This dual-path architecture ensures simultaneous fidelity to both the target signal distribution and the physical propagation environment.

\begin{algorithm}
\caption{RadioFlow Training Procedure}
\begin{algorithmic}[1]
\REQUIRE Target radio map $\mathbf{x}_1$, Gaussian noise $\mathbf{x}_0$, environmental context $\mathbf{c}$, model $M$, flow matcher $FM$, epochs $N$, learning rate $\eta$
\ENSURE Trained model $M$ for radio map synthesis
\STATE Initialize $M$ with random weights
\STATE Define optimizer $Opt$ (e.g., AdamW) with learning rate $\eta$
\STATE Initialize EMA model $M_{EMA}$ with decay $\gamma$ (e.g., 0.999)
\FOR {epoch = 1 to $N$}
    \FOR {batch in training data}
        \STATE Sample time $t \sim [0, 1]$ and compute conditional flow $\mathbf{u}_t = \mathbf{x}_1 - \mathbf{x}_0$
        \STATE Compute embedding $\mathbf{e} = M_{embed}(\mathbf{c})$ \COMMENT{Input conditioning}
        \STATE Predict vector field $\mathbf{v}_t = M(\mathbf{x}_t, t, \mathbf{e})$ \COMMENT{Flow-based generation}
        \STATE Calculate loss $L = \mathbb{E}\left[ \| \mathbf{v}_t - (\mathbf{x}_1 - \mathbf{x}_0) \|^2 \right]$
        \STATE Update $M$ using $Opt$ with gradient $\nabla L$
        \STATE Update $M_{EMA}$: $M_{EMA} \gets \gamma M_{EMA} + (1 - \gamma) M$
    \ENDFOR
    \STATE Adjust learning rate via scheduler (Warmup + Cosine)
    \IF {validation interval reached}
        \STATE Evaluate $M$ on validation set and plot loss
    \ENDIF
    \IF {save interval reached}
        \STATE Save $M$ and $M_{EMA}$ checkpoints
    \ENDIF
\ENDFOR
\STATE Refine output with spatial attention on $c$ \COMMENT{Contextual refinement}
\STATE \RETURN Trained $M$ and $M_{EMA}$
\end{algorithmic}
\end{algorithm}

The unified architecture achieves efficient radio map synthesis through its combination of flow-based direct transformation and environment-aware refinement, establishing a new paradigm for wireless channel modeling.

\subsection{Flow Matching for Radio Map Generation}
\label{subsec:flow_matching_radio}

Conditional Flow Matching (CFM) \cite{tong2024improvinggeneralizingflowbasedgenerative} represents a novel generative modeling paradigm designed for efficiently generating high-quality radio maps. Unlike conventional diffusion-based approaches that rely on iterative denoising through stochastic processes, CFM directly learns deterministic vector fields, enabling exact and rapid transport from Gaussian prior distributions to the target radio map distributions. This deterministic nature significantly simplifies the generation process, reducing computational overhead while preserving the detailed propagation characteristics essential in radio environments.

In the context of radio map generation, the model operates as a \textbf{conditional generative model}, where the generation process is explicitly guided by environmental context $\mathbf{c}$ (e.g., building layouts, terrain maps, transmitter locations). Formally, let $\mathbf{x}_0 \sim \mathcal{N}(\mathbf{0}, \mathbf{I})$ denote samples from a standard Gaussian prior and $\mathbf{x}_1 \sim q(\mathbf{x}_1|\mathbf{c})$ represent samples from the real radio map distribution conditioned on environment $\mathbf{c}$. CFM defines conditional probability paths by linear interpolation:
\begin{equation}
    \mathbf{x}_t = (1 - t)\mathbf{x}_0 + t\mathbf{x}_1 + \sigma_t\boldsymbol{\epsilon}, \quad \boldsymbol{\epsilon} \sim \mathcal{N}(\mathbf{0}, \mathbf{I}),
\end{equation}
where $\sigma \geq 0$ controls the interpolation noise level. The associated deterministic conditional vector field is simply defined as:
\begin{equation}
    \mathbf{u}_t(\mathbf{x}_0, \mathbf{x}_1) = \mathbf{x}_1 - \mathbf{x}_0.
\end{equation}
Note that the dependency of $\mathbf{u}_t$ on the environmental context $\mathbf{c}$ is implicit, arising from the fact that $\mathbf{x}_1$ is sampled from the conditional distribution $q(\mathbf{x}_1|\mathbf{c})$.

This concise yet expressive formulation allows for exact integration via ordinary differential equations (ODEs), dramatically reducing the inference complexity. Unlike iterative diffusion models \cite{cao2023surveygenerativediffusionmodel}, CFM's deterministic trajectories enable single-step or minimal-step sampling procedures, significantly lowering computational demand and making it particularly advantageous for latency-sensitive applications in 5G-Advanced and 6G networks \cite{9465788,9952510}. Additionally, CFM provides flexibility by supporting various probability interpolation strategies, such as optimal transport and Schrödinger bridge interpolations \cite{gushchin2023building,dimarino2019optimaltransportapproachschrodinger,GHOSAL2022109622}, thereby enhancing adaptability across diverse propagation scenarios.


\begin{table*}[t]
  \centering
  \setlength{\tabcolsep}{5pt}
  \begin{tabular}{l
                  S[table-format=1.4] S[table-format=2.2] S[table-format=1.4] S[table-format=1.4]
                  S[table-format=1.4] S[table-format=2.2] S[table-format=1.4] S[table-format=1.4]}
    \toprule
    \multirow{2}{*}{\textbf{Model}} &
    \multicolumn{4}{c}{\textbf{SRM}} &
    \multicolumn{4}{c}{\textbf{DRM}} \\
    \cmidrule(lr){2-5} \cmidrule(lr){6-9}
    & {NMSE$\downarrow$} & {PSNR$\uparrow$} & {RMSE$\downarrow$} & {SSIM$\uparrow$}
    & {NMSE$\downarrow$} & {PSNR$\uparrow$} & {RMSE$\downarrow$} & {SSIM$\uparrow$} \\
    \midrule
    RME-GAN        & 0.0115 & 30.54 & 0.0303 & 0.9323 & 0.0118 & 30.40 & 0.0307 & 0.9219 \\
    RadioUNet      & 0.0074 & 32.01 & 0.0244 & 0.9592 & 0.0089 & 31.75 & 0.0258 & 0.9410 \\
    UVM-Net        & 0.0085 & 30.34 & 0.0304 & 0.9320 & 0.0088 & 30.42 & 0.0301 & 0.9326 \\
    RadioDiff      & 0.0049 & 35.13 & 0.0190 & 0.9691 & 0.0057 & 34.92 & 0.0215 & 0.9536 \\
    RMDM           & 0.0031 & 34.47 & 0.0125 & \cellcolor{gray!20}{0.9780}
                   & 0.0047 & 34.21 & 0.0146 & \cellcolor{gray!20}{0.9680} \\
    \midrule
    RadioFlow Lite & 0.0077 & 34.92 & 0.0182 & 0.8897 & 0.0079 & 35.08 & 0.0178 & 0.8848 \\
    {RadioFlow Large}
                   & \cellcolor{gray!20}{0.0023} & \cellcolor{gray!20}{39.83} & \cellcolor{gray!20}{0.0103} & 0.9249
                   & \cellcolor{gray!20}{0.0028} & \cellcolor{gray!20}{39.37} & \cellcolor{gray!20}{0.0108} & 0.9236 \\
    \bottomrule
  \end{tabular}
  \caption{
    Quantitative comparison of representative models under two reconstruction paradigms: 
    SRM (Signal Recovery Module) and DRM (Denoising Reconstruction Module). 
    Metrics include NMSE, PSNR, RMSE, and SSIM. 
    \textbf{RadioFlow Large} achieves the best performance in NMSE, PSNR, and RMSE across both settings,
    while \textbf{RMDM} attains the highest SSIM in SRM and DRM.
  }
  \label{tab:srm_drm}
\end{table*}

\subsection{Training Strategy}

The training strategy for the Conditional Flow Matching (CFM) framework is designed to efficiently optimize a neural network for generating high-fidelity radio maps conditioned on environment-specific features, such as terrain elevation and urban morphology. During each training iteration, the model processes batches consisting of paired input features and target radio maps. For every sample in the batch, a latent variable $\mathbf{x}_0$ is drawn from a standard Gaussian distribution, and an interpolation time $t$ is sampled uniformly from the interval $[0,1]$. Based on these, an intermediate state $\mathbf{x}_t$ is constructed via interpolation, and the corresponding ground-truth conditional flow $\mathbf{u}_t$ is analytically derived.

The network is trained to predict the conditional vector field $\mathbf{v}_\theta(t, \mathbf{x}_t, \mathbf{c})$, with the objective of minimizing the mean squared error between the predicted and true flows. The training loss is formulated as:
\begin{equation}
    \mathcal{L}_{\text{CFM}}(\theta) = \mathbb{E}_{t, \mathbf{x}_0, \mathbf{x}_1, \mathbf{c}}\left[ \left\| \mathbf{v}_\theta(t, \mathbf{x}_t, \mathbf{c}) - (\mathbf{x}_1 - \mathbf{x}_0) \right\|_2^2 \right],
\end{equation}
ensuring that the model learns an accurate approximation of the underlying transport dynamics conditioned on the environmental context.

To further enhance controllability and generation quality, we employ \textbf{Classifier-Free Guidance (CFG)} \cite{ho2022classifierfreediffusionguidance}, a technique that enables flexible trade-offs between conditional fidelity and sample diversity. During training, the condition $\mathbf{c}$ is randomly dropped with probability $p_{\text{uncond}}$, forcing the model to learn both conditional and unconditional generation. At inference time, the predicted vector field is adjusted as $\mathbf{v}_{\theta}^{\text{guided}} = (1 + w) \mathbf{v}_\theta(t, \mathbf{x}_t, \mathbf{c}) - w \mathbf{v}_\theta(t, \mathbf{x}_t, \emptyset)$, where $w \geq 0$ is the guidance scale that controls the strength of conditioning.

To improve computational efficiency and numerical stability, automatic mixed-precision (AMP) training is employed, reducing memory consumption and accelerating convergence. Additionally, an exponential moving average (EMA) of the model parameters is maintained with a decay rate of 0.999 to provide a more stable representation of the model throughout training. A learning rate schedule combining an initial warm-up phase with cosine annealing \cite{10104453} is adopted to facilitate effective optimization and avoid premature convergence.

Model performance is periodically assessed on a held-out validation set. Training dynamics are monitored via loss curves, and model checkpoints, including both standard and EMA versions, are saved at regular intervals to ensure the preservation of optimal configurations. This comprehensive training scheme enables the proposed framework to achieve robust, stable, and highly accurate radio map generation, making it well-suited for real-time applications in next-generation wireless systems.

\section{Experiments}

To validate the effectiveness of RadioFlow, we conducted comprehensive experiments using the RadioMapSeer dataset, a specialized collection of simulated radio propagation data for urban environments. Our evaluation framework addresses three core aspects: (1) dataset characteristics, (2) static radio map (SRM) generation, and (3) dynamic radio map (DRM) modeling. The subsequent sections detail our experimental methodology and configuration parameters.

\begin{figure}[t]
    \centering
    \includegraphics[width=0.95 \linewidth]{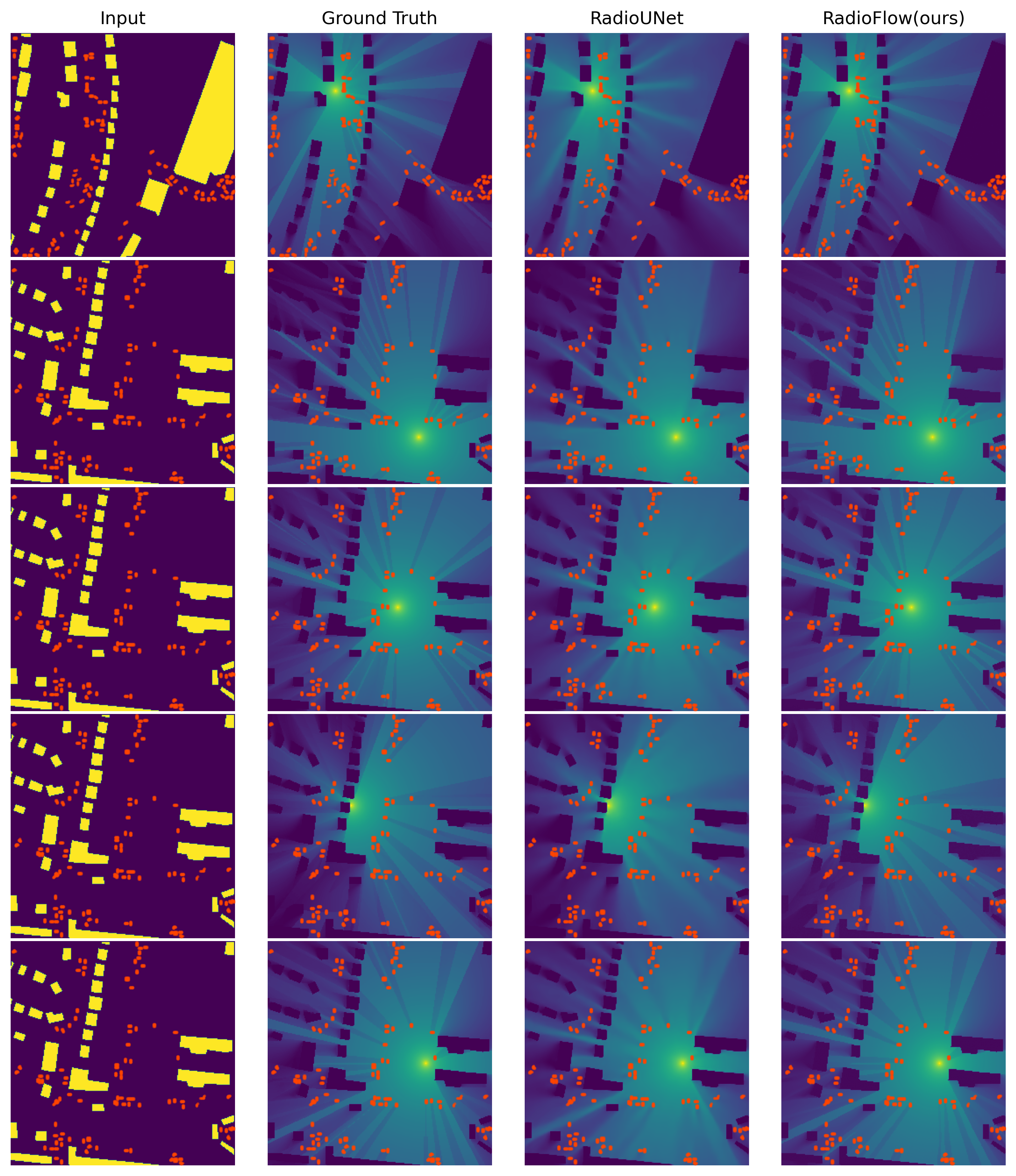}
    \caption{
        Visualization of the results generated by RadioFlow (DRM variant) compared to RadioUNet. 
        The proposed RadioFlow model demonstrates more fine-grained predictions and 
        exhibits closer alignment with the ground truth. 
    }
    \label{fig:drm_comparison}
\end{figure}

\begin{figure}[t]
    \centering
    \includegraphics[width=0.95 \linewidth]{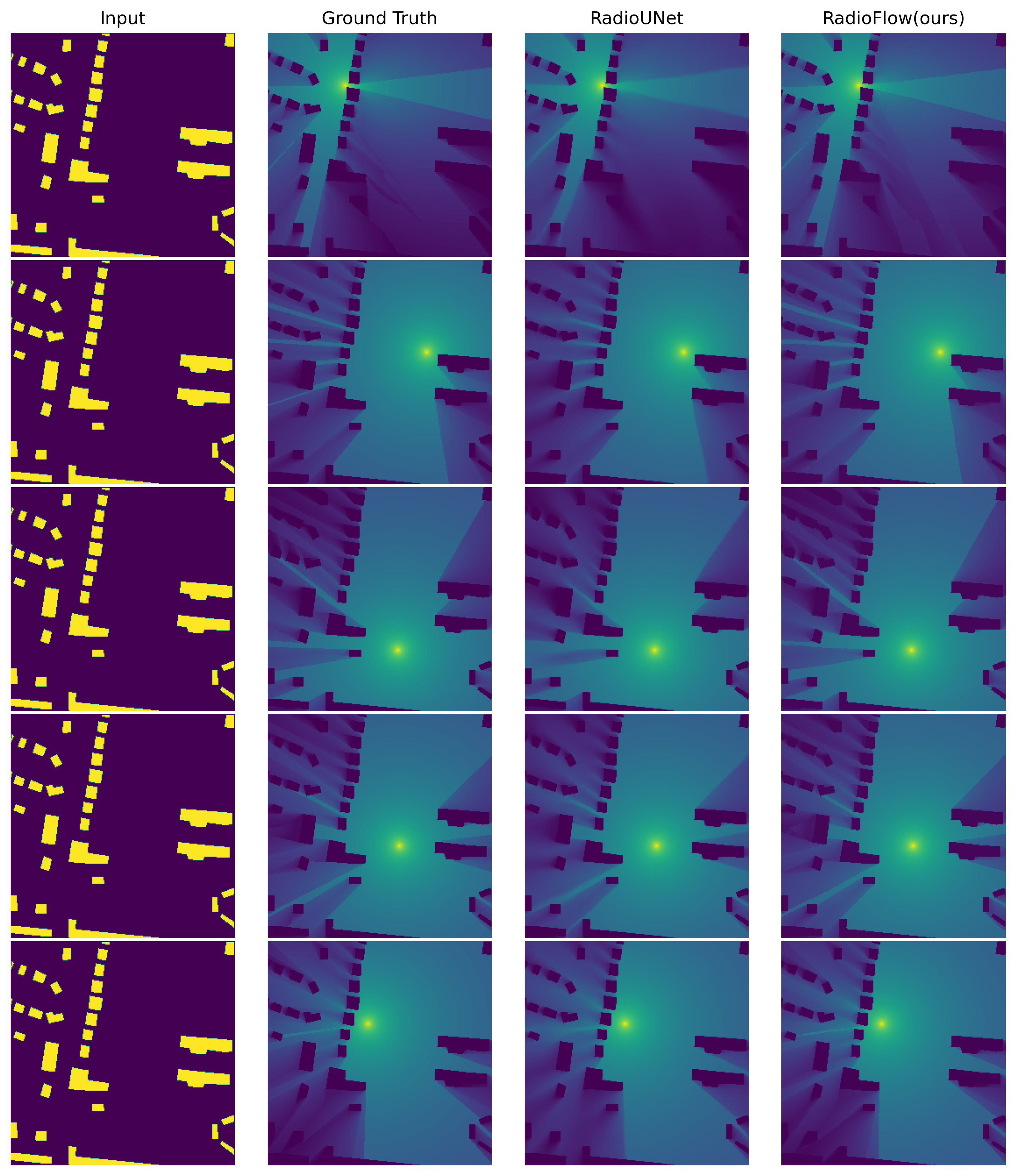}
    \caption{
        Visualization of the results generated by RadioFlow (SRM variant) compared to RadioUNet. 
        The SRM-based RadioFlow model provides visually coherent predictions and 
        captures more structural details in the signal distribution compared to RadioUNet.
    }
    \label{fig:srm_comparison}
\end{figure}

In addition to quantitative results, we present qualitative visual comparisons with the baseline RadioUNet. As shown in Fig.~\ref{fig:drm_comparison} (DRM) and Fig.~\ref{fig:srm_comparison} (SRM), RadioFlow produces more accurate and detailed signal maps, closely matching the ground truth and better reflecting spatial structures in both dynamic and static scenarios.

\subsection{Dataset Introduction}

The RadioMapSeer\cite{DatasetPaper} dataset provides a rigorous simulation platform \cite{7916282} for radio propagation studies, containing 701 distinct urban maps spanning six global cities (Ankara, Berlin, Glasgow, Ljubljana, London, and Tel Aviv). Each 256×256m urban area captures realistic architectural features extracted from OpenStreetMap data \cite{OpenStreetMap_PlanetDump_2017}, complete with 80 transmitter locations per map positioned at 1.5m height to simulate device-to-device communication scenarios.

The dataset's technical rigor derives from its hybrid simulation framework, integrating the Dominant Path Model (DPM) \cite{Wahl2005DominantPath} with Intelligent Ray Tracing (IRT) \cite{1040665} methodologies. These approaches enable precise modeling of complex signal propagation phenomena, encompassing reflections, diffractions, and penetration losses, across both static structures (e.g., buildings) and dynamic obstacles (e.g., vehicles). Pathloss values are computed at a 1-meter spatial resolution and archived as normalized PNG images. Dynamic elements are incorporated via procedurally generated vehicle placements (up to 100 vehicles per map) along road networks, enhancing the realism of the simulated environment.

The dataset is particularly suitable for our study due to the following key features:
\begin{itemize}[noitemsep, topsep=0pt, leftmargin=*]
    \item \textbf{Multiple ray-tracing complexity levels (IRT2/IRT4):} These enable sensitivity analysis of propagation effects.
    \item \textbf{Explicit separation of static and dynamic obstacles:} This permits controlled evaluation of environmental factors.
    \item \textbf{Global city sampling:} This ensures geographic diversity in urban morphology, ranging from the dense layouts of London to the broader avenues of Tel Aviv.
\end{itemize}
\subsection{Experiment Setup}

\paragraph{\textbf{Static Radio Map (SRM) Setup}}

For baseline SRM generation, we focus on time-invariant propagation effects caused by permanent structures. The model accepts two georeferenced inputs: a binary building map 
 (1 = structure, 0 = free space) and a transmitter location mask. The target output is a pathloss matrix where values quantify signal attenuation from building interactions.

Prior to training, we normalize all pathloss values to [0,1] using dataset-wide minimum/maximum thresholds. This normalization preserves relative signal strength patterns while improving neural network convergence. The 701 city maps are partitioned into 500 training and 201 testing instances, ensuring no geographic overlap between sets. Crucially, this split maintains consistent transmitter density (3.05 transmitters per 10,000m²) across both subsets to prevent sampling bias.

\paragraph{\textbf{Dynamic Radio Map (DRM) Setup}}

Extending the SRM framework, the DRM configuration introduces vehicular obstructions as dynamic elements. The input tensor now concatenates three channels: the static building map, transmitter mask, and a time-varying car distribution layer. Vehicles are modeled as 2×5×1.5m rectangular obstacles randomly placed along road segments, with density proportional to street capacity.

Pathloss normalization follows the SRM protocol, enabling direct comparison between static and dynamic scenarios. The training-test split remains identical to ensure urban layout consistency - only the presence/configuration of vehicles differs between SRM and DRM evaluations. This controlled variation isolates the impact of transient obstacles on propagation patterns.

The DRM task poses unique challenges: Vehicles introduce high-frequency spatial variations in pathloss fields while maintaining low temporal persistence. Our model must therefore distinguish between persistent building shadows and transient vehicular attenuations, requiring learned representations of obstacle mobility characteristics \cite{10.21203/rs.3.rs-3032515/v1}.

\subsection{Analysis}
We evaluate the performance of RadioFlow on both static radio map (SRM) and dynamic radio map (DRM) tasks, and compare it with several state-of-the-art baseline models. Table~\ref{tab:srm_drm} summarize the results for the SRM and DRM setups, respectively, while Table~\ref{tab:comparison} reports the average inference time and model complexity for each method.

\begin{table}[t]
    \centering
    \begin{tabular}{lcc}
        \hline
        \textbf{Method} & \textbf{Average time (s)} & \textbf{Parameters (M)} \\ \hline
       RME-GAN         & 0.042 &  10.61
       \\ 
        RadioUNet       & 0.056 & 9.71 \\ 
        RadioDiff       & 0.600 &  32.12
        \\ 
        RMDM            & 21.413                     & 82.83                 \\ \midrule
        RadioFlow       & 0.130                     & 52.53                  \\
        RadioFlow Lite  & 0.063                     & 3.89                   \\ \hline
    \end{tabular}
    \caption{
        Comparison of different methods in terms of inference efficiency and model complexity. RadioFlow achieves a good balance between speed and capacity, with significantly lower inference time than diffusion-based models (e.g., RadioDiff, RMDM) while maintaining strong performance. Notably, \textbf{RadioFlow Lite} further reduces both runtime and parameter count, making it suitable for resource-constrained deployment scenarios.
    }
    \label{tab:comparison}
\end{table}

\begin{table}[t]
\centering
\small
\setlength{\tabcolsep}{4pt}
\begin{tabular}{ccc|ccc}
\toprule
\textbf{Model} & \textbf{CFG} & ~ & NMSE$\downarrow$ & PSNR$\uparrow$ & RMSE$\downarrow$ \\
\midrule
Full & 1.0 & & 0.0024 & 39.83 & 0.0103 \\
{Full} & {1.5} & & \cellcolor{gray!20}{0.0023} & \cellcolor{gray!20}{39.83} & \cellcolor{gray!20}{0.0103} \\
Full & 2.0 & & 0.0025 & 39.65 & 0.0105 \\
Full & 2.5 & & 0.0026 & 39.45 & 0.0109 \\
Full & 3.0 & & 0.0029 & 38.94 & 0.0114 \\
Full & 3.5 & & 0.0032 & 38.54 & 0.0119 \\
Full & 4.0 & & 0.0036 & 38.14 & 0.0125 \\
Full & 4.5 & & 0.0039 & 37.74 & 0.0131 \\
Full & 5.0 & & 0.0043 & 37.33 & 0.0138 \\
Full & 5.5 & & 0.0047 & 36.98 & 0.0143 \\
Full & 6.0 & & 0.0051 & 36.60 & 0.0150 \\
\midrule
Lite & 0.0 & & 0.0077 & 34.92 & 0.0182 \\
\bottomrule
\end{tabular}
\caption{
Ablation study of Classifier-Free Guidance (CFG) in the RadioFlow model under the SRM setting. 
Best performance is achieved at CFG = 1.5. Larger CFG degrades performance, and the Lite version (CFG = 0) performs worst.
}
\label{tab:cfg_srm}
\end{table}

\begin{table}[t]
\centering
\small
\setlength{\tabcolsep}{4pt}
\begin{tabular}{ccc|ccc}
\toprule
\textbf{Model} & \textbf{CFG} & ~ & NMSE$\downarrow$ & PSNR$\uparrow$ & RMSE$\downarrow$ \\
\midrule
Full & 1.0 & & 0.0029 & 39.23 & 0.0110 \\
{Full} & {1.5} & & \cellcolor{gray!20}{0.0028} & \cellcolor{gray!20}{39.37} & \cellcolor{gray!20}{0.0108} \\
Full & 2.0 & & 0.0029 & 39.23 & 0.0110 \\
Full & 2.5 & & 0.0031 & 38.94 & 0.0114 \\
Full & 3.0 & & 0.0034 & 38.53 & 0.0119 \\
Full & 3.5 & & 0.0037 & 38.11 & 0.0125 \\
Full & 4.0 & & 0.0042 & 37.67 & 0.0132 \\
Full & 4.5 & & 0.0046 & 37.22 & 0.0139 \\
Full & 5.0 & & 0.0051 & 36.82 & 0.0145 \\
Full & 5.5 & & 0.0056 & 36.42 & 0.0152 \\
Full & 6.0 & & 0.0061 & 36.05 & 0.0159 \\
\midrule
Lite & 0.0 & & 0.0079 & 35.08 & 0.0178 \\
\bottomrule
\end{tabular}
\caption{
Ablation study of Classifier-Free Guidance (CFG) in the RadioFlow model under the DRM setting.
CFG = 1.5 provides the best performance, while both higher CFG values and the Lite variant result in worse accuracy.
}
\label{tab:cfg_drm}
\end{table}

        

\paragraph{\textbf{SRM Performance.}}
As shown in Table~\ref{tab:srm_drm}, RadioFlow achieves a PSNR of 39.83 and an NMSE of 0.0023, outperforming all baselines including the diffusion-based RadioDiff and the physics-informed RMDM. Although RMDM attains a comparable NMSE of 0.0031, its improvement relies on substantially higher architectural complexity and a larger number of parameters. In contrast, RadioFlow maintains high reconstruction fidelity with remarkable efficiency, demonstrating its capability to preserve spatial structures in static propagation environments.

\paragraph{\textbf{DRM Performance.}}
Table~\ref{tab:srm_drm} highlights the advantages of RadioFlow in dynamic environments, where channel characteristics vary due to moving obstacles (e.g., vehicles). RadioFlow achieves the best overall performance with an NMSE of 0.0028 and a PSNR of 39.37, clearly surpassing RadioDiff (NMSE: 0.0057) and RMDM (NMSE: 0.0047). These results confirm the robustness of our approach in modeling spatio-temporal variations, which is crucial for next-generation real-time wireless sensing and communication applications.

\paragraph{\textbf{Model Efficiency.}}
We also compare the computational efficiency of different methods in Table~\ref{tab:comparison}. Although RadioDiff achieves competitive accuracy, it requires the longest inference time (0.60\,s) and a relatively large parameter count (32.12\,M). In contrast, RadioFlow achieves a much lower inference time of 0.13\,s while maintaining strong performance, demonstrating an effective trade-off between accuracy and latency. The lightweight variant, RadioFlow~Lite, further reduces the model size to just 3.89\,M parameters with an inference time of 0.063\,s, making it suitable for deployment in resource-constrained or edge scenarios.

In summary, RadioFlow consistently outperforms state-of-the-art baselines on both static and dynamic setups while offering up to \textbf{4--5$\times$ faster inference} and a significantly smaller model footprint. These results confirm the effectiveness of the proposed flow-based generative paradigm for real-time and accurate radio map generation.

\subsection{Ablation Study}

We conduct extensive ablation studies to evaluate the effectiveness of key components and hyperparameters in our proposed \textbf{RadioFlow} framework. The following aspects are analyzed: (1) the influence of Classifier-Free Guidance (CFG) scale, (2) the role of architectural modules such as Exponential Moving Average (EMA) and Spatial Attention (SA), and (3) the number of \textbf{ODE integration steps} during inference.

\paragraph{\textbf{Effect of CFG Scale.}}
Tables~\ref{tab:cfg_srm} and \ref{tab:cfg_drm} report performance under varying CFG scales for both SRM and DRM settings. We observe that moderate CFG scales (particularly 1.5) yield the best performance across all metrics (NMSE, PSNR, RMSE), indicating a balanced trade-off between conditional fidelity and generative diversity. As the CFG scale increases further, performance degrades consistently, suggesting that excessive guidance leads to over-regularization. Conversely, CFG = 0 (i.e., no classifier guidance, or the Lite variant) performs the worst, highlighting the necessity of guidance-based control.

\paragraph{\textbf{Impact of EMA and Spatial Attention.}}
To evaluate the contribution of architectural components, we ablate the use of EMA and SA, as shown in Table~\ref{tab:ablation}. Removing either component leads to noticeable performance drops in both SRM and DRM configurations. Disabling SA introduces the largest error, emphasizing its effectiveness in capturing spatial dependencies for robust signal reconstruction. The full model with both EMA and SA consistently achieves the lowest NMSE and RMSE and the highest PSNR.

\paragraph{\textbf{Effect of ODE Integration Steps.}}
Unlike traditional diffusion models that require many timesteps, RadioFlow performs deterministic integration following the flow-matching principle. The number of ODE integration steps reflects the trade-off between computational cost and reconstruction accuracy. As shown in Table~\ref{tab:timesteps_ablation}, the best performance is obtained with a single integration step, suggesting that RadioFlow can reconstruct high-fidelity maps efficiently. Increasing the number of steps results in negligible improvements and even slight degradation, likely due to over-smoothing effects. These results indicate that RadioFlow is both computationally efficient and robust under short-step inference.

Overall, the ablation results validate the importance of each proposed component. CFG tuning is critical for balancing precision and generalization, while both EMA and SA modules substantially enhance performance. The model is also computation-efficient, achieving strong results even with a single-step ODE integration.

\begin{figure}[t]
    \centering
    \includegraphics[width=0.95 \linewidth]{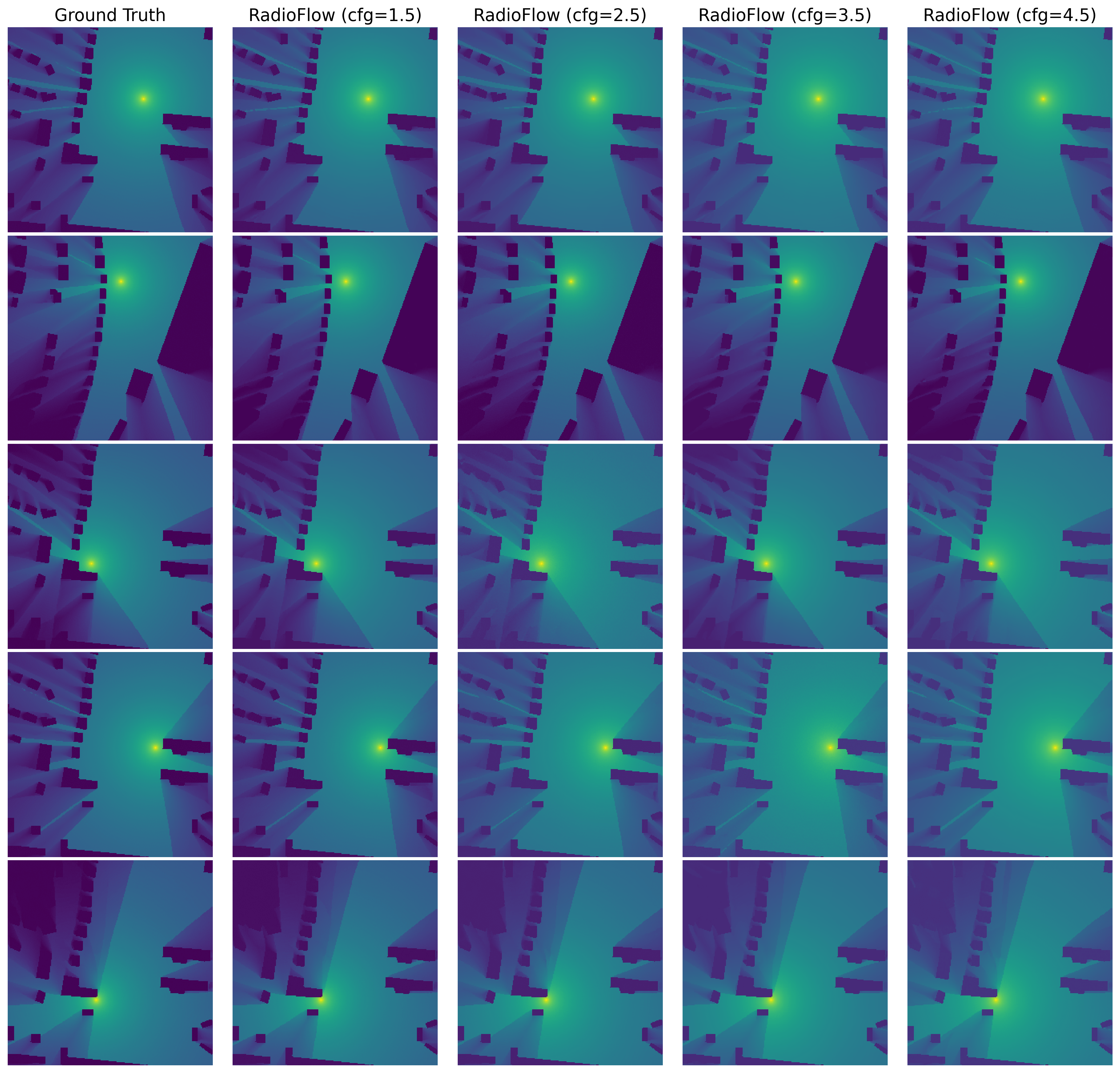}
    \caption{Comparison of SRM results obtained using different cfg scales in the RadioFlow model. The cfg scale values used in the experiment are 1.5, 2.5, 3.5, 4.5, and 5.5. As shown, varying the cfg scale affects the model's output, with higher cfg values generally yielding more detailed and sharper results. These results are compared to the ground truth to illustrate the impact of cfg scaling on model performance.}
    \label{fig:srm_cfg_comparison}
\end{figure}

\begin{table}[t]
  \centering
  \label{tab:ablation}
  \setlength{\tabcolsep}{6pt} 
  \begin{tabular}{ll
                  S[table-format=1.4,table-number-alignment=center]
                  S[table-format=2.2,table-number-alignment=center]
                  S[table-format=1.4,table-number-alignment=center]}
    \toprule
    \multirow{2}{*}{Setup} & \multirow{2}{*}{Method} &
    \multicolumn{3}{c}{\textbf{Metrics}} \\ \cmidrule(lr){3-5}
    & & {NMSE\,$\downarrow$} & {PSNR\,$\uparrow$} & {RMSE\,$\downarrow$} \\
    \midrule
    \multirow{3}{*}{DRM} 
        & w/o EMA                       & 0.0033 & 38.67 & 0.0117 \\
        & w/o SA                & 0.0038 & 38.07 & 0.0160 \\ 
        & w/ EMA + SA                   & \cellcolor{gray!20}\bfseries 0.0029 & \cellcolor{gray!20}\bfseries 39.24 & \cellcolor{gray!20}\bfseries 0.0110 \\
    \midrule
    \multirow{3}{*}{SRM} 
        & w/o EMA                       & 0.0035 & 38.55 & 0.0167 \\
        & w/o SA                & 0.0047 & 37.06 & 0.0106 \\ 
        & w/ EMA + SA                   & \cellcolor{gray!20}\bfseries 0.0023 & \cellcolor{gray!20}\bfseries 39.83 & \cellcolor{gray!20}\bfseries 0.0103 \\
        
    \bottomrule
  \end{tabular}
  \caption{Ablation study evaluating the impact of Exponential Moving Average (EMA) and Spatial Attention (SA). The presented results were identical for both the DRM and SRM experimental setups. Lower Normalized Mean Square Error (NMSE) / Root Mean Square Error (RMSE) and higher Peak Signal-to-Noise Ratio (PSNR) indicate better performance. Best results achieved are highlighted in \textbf{bold}.}
  \label{tab:ablation}
  \vspace{-0.5em}
\end{table}

\begin{table}[t]
\centering
\small
\begin{tabular}{c|cccc}
\toprule
\textbf{Timesteps} & \textbf{NMSE} $\downarrow$ & \textbf{PSNR} $\uparrow$ & \textbf{RMSE} $\downarrow$ \\
\midrule
1   & \cellcolor{gray!20}0.0079 & \cellcolor{gray!20}35.08 & \cellcolor{gray!20}0.0178 \\
5   & 0.0084 & 34.78 & 0.0184 \\
10  & 0.0084 & 34.79 & 0.0184 \\
20  & 0.0084 & 34.79 & 0.0184 \\
50  & 0.0084 & 34.78 & 0.0184 \\
\bottomrule
\end{tabular}
\caption{
Ablation results evaluating the impact of different timesteps on the performance of the model. The table presents the NMSE, PSNR, and RMSE metrics for various timesteps, where smaller NMSE and RMSE values and higher PSNR values indicate better performance. The best performance is observed at timestep 1, and performance tends to stabilize with higher timesteps (5, 10, 20, 50). These results suggest that increasing the number of timesteps does not significantly improve performance beyond a certain point. 
}
\label{tab:timesteps_ablation}
\end{table}
\section{Conclusion}

This paper introduces \textbf{RadioFlow}, a novel generative framework leveraging Flow Matching with Continuous Normalizing Flows (CNFs) for efficient, high-fidelity radio map generation. Unlike diffusion-based methods requiring computationally intensive iterative denoising, RadioFlow employs deterministic conditional vector fields to achieve single-step synthesis, significantly reducing complexity while maintaining accuracy.

Evaluations on the RadioMapSeer dataset show that RadioFlow outperforms state-of-the-art baselines, including RadioUNet, RadioDiff, and RMDM, across static (SRM) and dynamic (DRM) tasks. It exhibits superior accuracy in dynamic scenarios, underscoring its robustness and suitability for real-time 5G-Advanced and 6G network applications.

Ablation studies confirm the efficacy of key components—Classifier-Free Guidance (CFG), Exponential Moving Average (EMA), and Spatial Attention (SA). Optimal CFG balances fidelity and generalization, while EMA and SA enhance stability and spatial coherence in map generation.

Additionally, \textbf{RadioFlow Lite}, a compact variant, delivers competitive performance with reduced inference time and parameters, ideal for resource-constrained edge deployments.

In summary, RadioFlow provides a scalable, efficient solution for next-generation radio map construction. Its single-step generation and architectural simplicity advance wireless channel modeling, facilitating intelligent, real-time network optimization for future 6G systems.

\bibliographystyle{IEEEtran}
\bibliography{main}

\end{document}